\documentclass[sigconf]{acmart}

\usepackage{booktabs} 
\pdfoutput=1
\usepackage{times}  
\usepackage{helvet}  
\usepackage{courier}  
\usepackage{graphicx}  
\usepackage{mathtools}
\usepackage{booktabs}
\usepackage{color}
\usepackage{hyperref}
\usepackage{dblfloatfix}
\usepackage{subfig}
\usepackage{etoolbox}
\usepackage{calc}
\usepackage{xcolor}
\usepackage{algorithmic}
\usepackage[ruled,vlined]{algorithm2e}
\usepackage{mdwlist}
\setcopyright{rightsretained}

\usepackage{xspace}
\usepackage{latexsym}
\usepackage{amsfonts}
\usepackage{amsmath}
\usepackage{amssymb}
\usepackage{color}
\usepackage{colortbl}
\usepackage{epsfig}
\usepackage{graphicx}
\usepackage{paralist}
\usepackage{enumerate}
\usepackage[color,matrix,arrow,all]{xy}
\usepackage{comment}
\usepackage{booktabs}
\usepackage{balance}
\usepackage{stmaryrd}
\usepackage{pifont}
\usepackage{hhline}
\usepackage{listings}
\usepackage{array}
\usepackage{float}
\usepackage[flushleft]{threeparttable}
\usepackage{graphicx,wrapfig,lipsum}

\DeclareMathAlphabet{\pazocal}{OMS}{zplm}{m}{n}

\usepackage{array}
\newcolumntype{M}{>{\begin{varwidth}{2cm}}l<{\end{varwidth}}}
\newcolumntype{L}[1]{>{\raggedright\let\newline\\\arraybackslash\hspace{0pt}}m{#1}}
\newcolumntype{C}[1]{>{\centering\let\newline\\\arraybackslash\hspace{0pt}}m{#1}}
\newcolumntype{R}[1]{>{\raggedleft\let\newline\\\arraybackslash\hspace{0pt}}m{#1}}

\newcommand{\ie}{{\em i.e.,}\xspace}
\newcommand{\eg}{{\em e.g.,}\xspace}

\newcommand{\aka}{\emph{a.k.a.}\xspace}

\newenvironment{noindlist}
 {\begin{list}{\labelitemi}{\leftmargin=0em \itemindent=1em}}
 {\end{list}}

\newcommand{\Ni}{({\em i})~}
\newcommand{\Nii}{({\em ii})~}
\newcommand{\Niii}{({\em iii})~}

\newcommand{\Niv}{({\em iv})~}
\newcommand{\Nv}{({\em v})~}

\newcommand{\acttovec}{\texttt{act2vec}}
\newcommand{\sampletovec}{\texttt{sample2vec}}
\newcommand{\hourtovec}{\texttt{hour2vec}}
\newcommand{\daytovec}{\texttt{day2vec}}
\newcommand{\weektovec}{\texttt{week2vec}}
\newcommand{\doctovec}{\texttt{doc2vec}}
\newcommand{\wordtovec}{\texttt{word2vec}}

\newcommand{\sampletovecreg}{\texttt{sample2vec+Reg}}
\newcommand{\hourtovecreg}{\texttt{hour2vec+Reg}}
\newcommand{\daytovecreg}{\texttt{day2vec+Reg}}

\DeclareMathOperator \real{\mathbb{R}}

\newcommand{\Ls}{\mathcal{L}}

\newcommand{\Ns}{\pazocal{N}}
\newcommand{\Vs}{\pazocal{V}}

\newcommand{\Us}{\pazocal{U}}

\newcommand{\Ds}{\pazocal{D}}

\newcommand{\ph}{\mathbf{\Phi}}
\newcommand\norm[1]{\left\lVert#1\right\rVert}

\acmConference[WOODSTOCK'97]{ACM Woodstock conference}{July 1997}{El
  Paso, Texas USA} 
\acmYear{1997}
\copyrightyear{2016}

\begin{document}
\title[Co-Morbidity Exploration on Wearables Activity Data]{Co-Morbidity Exploration on Wearables Activity Data Using Unsupervised Pre-training and Multi-Task Learning}

\author{Karan Aggarwal}
\affiliation{%
  \institution{University of Minnesota}
}
\email{aggar081@umn.edu}

\author{Shafiq Joty}
\affiliation{%
  \institution{Nanyang Technological University}
}
\email{srjoty@ntu.edu.sg}
 
\author{Luis F. Luque}
\affiliation{%
  \institution{QCRI}
  }
\email{lluque@hbku.edu.qa}

\author{Jaideep Srivastava}
\affiliation{%
  \institution{University of Minnesota}
}
\email{srivasta@umn.edu}

\begin{abstract}

Physical activity and sleep play a major role in the prevention and management of many chronic conditions. It is not a trivial task to  understand their impact on chronic conditions. Currently, data from electronic health records (EHRs), sleep lab studies, and activity/sleep logs are used. The rapid increase in the popularity of wearable health devices provides a significant new data source,  making it possible to track the user's lifestyle real-time through web interfaces, both to consumer as well as their healthcare provider, potentially. However, at present there is a gap between lifestyle data (\eg\ sleep, physical activity) and clinical outcomes normally captured in EHRs. This is a critical barrier for the use of this new source of signal for healthcare decision making. Applying deep learning to wearables data provides a new opportunity to overcome this barrier.
 
To address the problem of the unavailability of clinical data from a major fraction of subjects and unrepresentative subject populations, we propose a novel unsupervised (task-agnostic) time-series representation learning technique called \acttovec. \acttovec~ learns useful features by taking into account the co-occurrence of activity levels along with periodicity of human activity patterns. The learned representations are then exploited to boost the performance of disorder-specific supervised learning models. Furthermore, since many disorders are often related to each other, a phenomenon referred to as co-morbidity, we use a multi-task learning framework for exploiting the shared structure of disorder inducing life-style choices partially captured in the wearables data. Empirical evaluation using 28,868 days of actigraphy data from 4,124 subjects shows that our proposed method performs and generalizes substantially better than the conventional time-series symbolic representational methods and  task-specific deep learning models.
 
\end{abstract}

%
%

\maketitle

\section{Introduction}
\label{sec:intro}


Physical activity and sleep are crucial to human wellbeing. The benefits of physical activity and sleep are paramount, including prevention of physical and cognitive disorders such as cancer or diabetes~\cite{warburton2006health}. Sleep deprivation and poor physical activity habits severely impact quality of life ~\cite{mcclain2014associations}. The current rise in chronic conditions, mainly due to aging and unhealthy lifestyles, is putting our healthcare systems under stress with long waiting times leading to delays in diagnosis of health disorders. For sleep-related disorders, the economic cost of those delays is enormous~\cite{shelgikar2014multidisciplinary}, with a major sleep disorder, obstructive sleep apnea syndrome, alone costing \$87 billion per year~\cite{apnea,watson2016health} of estimated productivity loss in USA. 




In order to study sleep problems, subjects have to go through different diagnosing steps, often involving \emph{polysomnography} (PSG) studies which can require  an overnight stay in the lab. Traditionally, health professionals need to relay on patient subjective feedback to understand health behaviors such as sleep or physical activity in the real world. Problems with recall and subjectivity have raised interest on using wearables to better study sleep and physical activity. That potential is now growing with the increasing popularity of health and fitness wearables. We now have an ability to track a subject's physical activity and sleep patterns in real-time through online data vaults like \textit{Google Fit}, providing access alike to consumers and health-care providers. Often, these connected devices are integrated in an ecosystem where data like weight and blood pressure is also available, thanks to other consumer health devices. The wearables market hit \$14 billion in 2016, and is expected to rise to \$34 billion by 2020 \cite{wmarket}. With over 411 million expected shipments in 2020, a significant proportion of the population, at least in the high income countries, can be expected to possess wearables. 

An automated tracking system that collects human activity signals from wearable devices in real-time, mines the activity patterns to extract useful relevant information, can go a long way for health-care delivery. Such system can reduce the waiting times by helping in identifying subjects at risk, monitor their compliance during therapy, and provide real-time recommendations based on consumer's behavior~\cite{yom2017encouraging}. This has the potential to provide significant savings in health-care costs, and improved lifestyle due to early detection of potentially debilitating conditions. 



Although analysis of wearables data for the diagnosis of health problems has significant benefits, a major challenge is the availability of EHR data for only a (very small) fraction of subjects who consent research surveys or studies. Not only does it render useless the activity data from majority of subjects, it might give an unrepresentative sample of disorder-positive population with respect to the general population. Hence, any approach towards using physical activity signals should be designed to take into account the generalization of the approach. Task-specific supervised learning tends to generalize poorly with skewed datasets. This challenge is exacerbated by the noisy nature of activity signals, and small dataset size of subjects who underwent diagnosis. 


Traditional time-series analysis use symbolic representation like  Symbolic Aggregate Approximation (SAX)~\cite{lin2007experiencing, senin2013sax} converting time-series into a symbolic sequence by assuming a distribution over the symbols. 
Despite successful applications in a wide variety of tasks involving classification and clustering \cite{bagnall2017great}, the symbolic representation methods are limiting in several ways. First,  
it prevents the model from considering sufficiently long sequential dependencies, leading to the so-called \emph{curse of dimensionality} problem \cite{Bengio03}. 
As a result, traditional methods use bag-of-words (BoW) representations like TF-IDF vectors. Second, due to its high dimensions, the traditional vector space models often suffer from  sparsity problems, making the prediction model inefficient \cite{le2014distributed}.  




Our contributions in this paper address the above-mentioned challenges and remedy the problems of symbolic representations. We conduct our research in three main steps as outlined below: 

\begin{enumerate}[(a)]
\item \textbf{Learn disorder-agnostic representation (embeddings) for activity signals:} To utilize the large amounts of unlabeled human activity data, we propose a task-agnostic (unsupervised) representation learning method \acttovec\ that learns \emph{condensed} vector representation for time-series activity signals from raw activity data (\ie\ without using diagnosis information). \acttovec\ uncovers the common patterns of human activity by means of \emph{distributed} representation, which can then be leveraged towards diagnosis prediction tasks. One of the long standing challenges in the time-series domain is the selection of granularity for time-segments (\ie\ time windows), which serve as the basic analysis units. We explore learning representations at various levels of time granularity, spanning over 30-seconds (device rate), an hour, a day, and a week. We devise a novel learning algorithm that optimizes two different measures to capture local and global patterns in a time-series along with a smoothing criteria. 

\item \textbf{Boost disorder-specific supervised learning using pre-trained embeddings:} Since the embeddings are learned from a large dataset of human activity signals, they capture distributional similarity between the signal levels, and are known to generalize well across tasks. It has been shown that adding unsupervised pre-trained vectors to initialize the supervised models produces better performance~\cite{krizhevsky2012imagenet,collobert2011natural,hinton2012deep}. 
Following this trend, we use pre-trained \acttovec\ embeddings to boost the performance of our supervised disorder prediction models that are based on \emph{convolution neural networks}.

\item \textbf{Exploit co-morbidity with multi-task learning:} Co-morbidity is a common phenomenon in medicine that indicates, presence of a disorder can cause (or can be caused by) another disorder in the same patient, \ie\ disorders can be co-related \cite{valderas2009defining}. In this paper, we propose a \emph{multi-task deep learning framework} to utilize co-morbidity. The framework captures dependencies between multiple random variables representing disorders, and promotes generalization by inducing features that are informative for multiple disorder prediction tasks. Co-morbidity  has been successfully exploited previously in different settings, \eg\  for clinical visits \cite{razavian2016multi} and clinical diagnosis \cite{lipton2015learning}.



\end{enumerate}



We use two publically available health wearable (actigraphy) datasets~\cite{sorlie2010design,bild2002multi} for training our model on 28,868 days of actigraphy data across 4,124 subjects. We evaluate our approach against existing models and baselines on four disorder prediction tasks -- Sleep Apnea, Diabetes, Hypertension, and Insomnia. Our main findings are the following:

\begin{enumerate}[(i)]
\item Our proposed \acttovec\ representation learning method (using linear classifier) outperforms existing time-series symbolic representation vector space models with a good margin, with day level representations performing the best;

\item The pre-trained embeddings from \acttovec\ improve performance of the supervised learning methods with task-specific as well as multi-task objectives; and

\item Using a multi-task learning approach helps exploit co-morbidity, \emph{boosting} the performance over individual supervised disorder prediction tasks. 
\end{enumerate}


The remainder of this article is organized as follows. After reviewing related work in Section \ref{sec:rel}, we define the problem formally in Section \ref{sec:problem}. In Section~\ref{sec:model}, we present our complete deep learning framework comprising our representation learning model \acttovec\ (section \ref{subsec:acttovec}), our CNN model as the supervised prediction model (section \ref{subsec:cnn}), and the multi-task learning framework (section \ref{subsec:multitask}). After describing experimental settings in Section \ref{sec:settings}, we present our results and analysis in Section \ref{sec:results}. Finally, we conclude with future directions in Section \ref{sec:conclusion}.

\section{Related Work} \label{sec:rel}

\begin{figure*}[t!]
  \centering
  \includegraphics[scale=0.48]{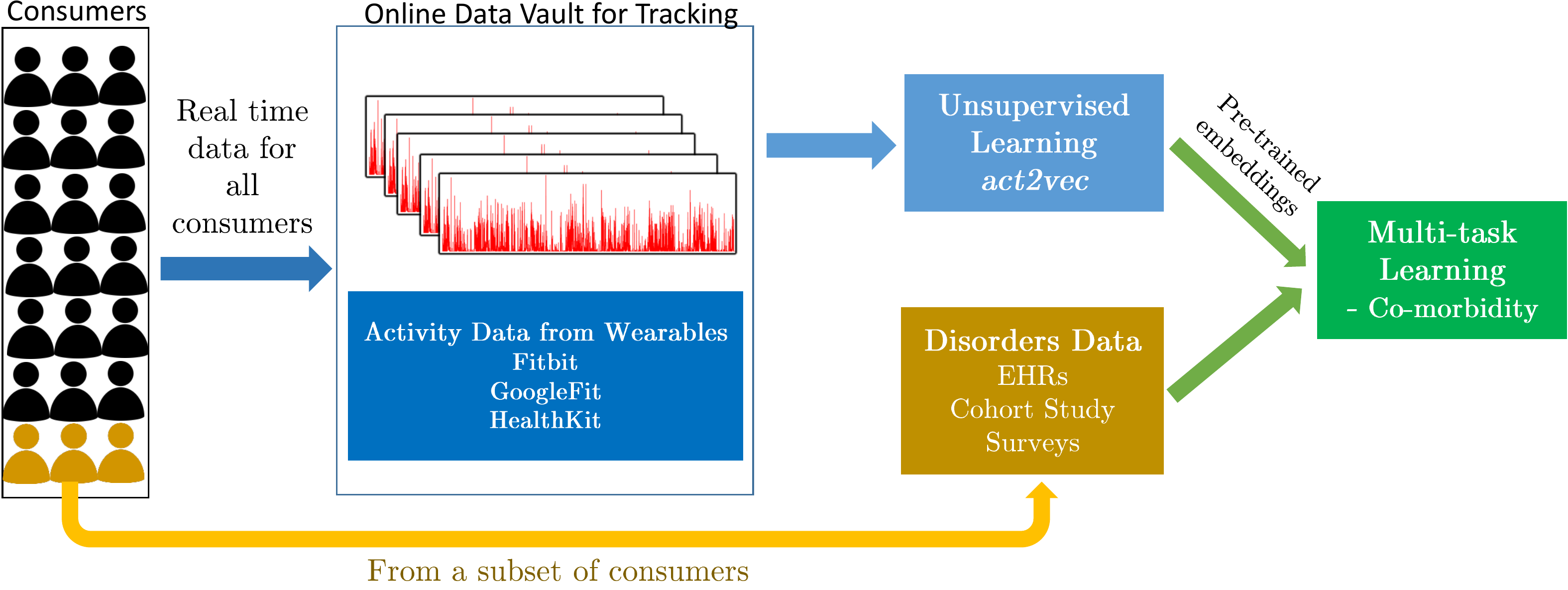}
  \caption{Work-flow of our proposed solution. Consumers track their activity using online data vaults like Google Fit, with a fraction of consumers' diagnosis data available through survey or EHR consent. Our proposed methods use both data sources for improved disorder prediction. }
  \label{fig:workflow}
\end{figure*}

We divide related works in four parts as described briefly below: \Ni human 
activity recognition, \Nii representational learning, \Niii time-series analysis methods, and \Niv co-morbidity literature.
\paragraph{Human activity research} Human activity has been a widely studied area 
especially the problem of human activity recognition (HAR) with the goal of recognizing human activity from
a stream of data such as camera recordings, motion 
detectors, and accelerometers. Wearable sensors like 
accelerometers (actigraphy) have mostly been used for human activity recognition task 
in machine learning~\cite{bulling2014tutorial,alsheikh2016deep}, while 
medical practitioners perform manual examination on the actigraphy 
data for diagnosing mostly sleep-disorders~\cite{sadeh2011role}. 
Recent works~\cite{sathyanarayana2016impact} have tried using actigraphy data for quantifying sleep quality using deep learning. The main difference with our method being that we present task-agnostic and generalizable models rather than plain end-to-end learning. With connected devices data, actigraphy is being deployed 
as auxiliary  to actively monitor human behavioral patterns with an 
aim for real-time monitoring~\cite{yom2017encouraging,althoff2017harnessing}.

\paragraph{Representation Learning}
\citeauthor{bengio2013representation}~\cite{bengio2013representation} provide an overview of 
representation learning that is used to learn good features from the raw input 
space that are powerfully discriminative for downstream tasks. 
It is based on ideas of better network convergence by adding (unsupervised) pre-trained vectors and better encoding of mutual information of input features 
at the input layer~\cite{goodfellow2016}. In past 
couple of years, the area has made enormous progress in natural language 
processing~\cite{collobert2011natural}, computer 
vision~\cite{krizhevsky2012imagenet}, and speech 
recognition~\cite{hinton2012deep}. Of particular interest 
are the developments in natural language processing with distributed bag-of-words (DBOW) architectures~\cite{mikolov2013distributed} optimized to predict the 
context of the language unit (\eg word) at hand, unlike continuous-bag-of-
words (CBOW) that predicts the language unit from its context. The DBOW model has been extended to incorporate discourse context \cite{saha-joty-hasan-ecml-17} and the node embeddings in networks \cite{Grover.Leskovec:16}. In a similar fashion, we use 
DBOW to capture local patterns in a time segment.

\paragraph{Time series analysis methods} 
Time series methods use pair-wise similarity concept to perform 
classification~\cite{bagnall2017great} and clustering tasks, with euclidean distance as the measure of 
similarity.  Dynamic Time Warping~\cite{berndt1994using} is a widely used 
technique for finding similarity between two time-series with totally 
different basal time units. However, it is extremely computationally 
expensive and its pair-wise similarity approach renders it non-scalable.
This has lead to creation of time-series symbolic representation techniques
like SAX (Symbolic Aggregate Approximation)~\cite{lin2007experiencing}, that 
convert time-series into a symbolic sequence that can be further used for 
feature extraction. SAX-VSM (SAX-vector space model) uses tf-idf (term 
frequency-inverse document frequency) transformation of these symbolic 
sequences to get vector representation of sequence windows. 
BOSS~\cite{schafer2015boss} is a symbolic representation technique that uses 
Fourier transform of the time-series-windows to create symbolic sequences. 
BOSS-VS~\cite{schafer2016scalable} creates vector space  
in a similar fashion to SAX-VSM. 

\paragraph{Co-morbidity}  A number of studies in the health informatics have 
exploited co-morbidity using multi-task learning~\cite{berndt1994using} for 
improving diagnosis~\cite{wang2014multi,che2017boosting,che2017boosting,caruana1996using}.
Co-morbidity structures have been used for predicting clinical 
visits~\cite{razavian2016multi} and diagnosis~\cite{lipton2015learning}.
While activity data from wearables has long been used for sleep related 
disorders, even cancer~\cite{fitcancer}, exploiting co-morbidity on activity 
signals has been relatively unexplored. Our multi-task framework is closest 
to Collobert et al.~\cite{collobert2011natural}'s convolutional
neural network for multi-task learning. In the next section, we describe the problem and opportunities posed.

\begin{figure}[t!]
  \centering
  \includegraphics[width=0.47\textwidth]{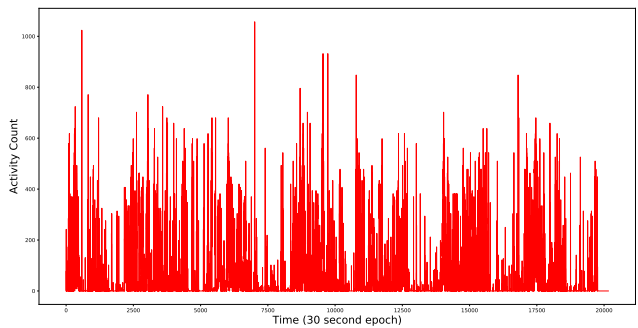}
  \vspace{-0.5em}
  \caption{A user's activity time-series over a week.}
  \label{fig:activity}
\end{figure}

\section{Problem Statement} \label{sec:problem}

Wearables data has created new opportunities to understand how human behaviors such as physical activity and sleep affect our cognitive and physical health. This is especially important given that there are millions of users of wearable devices.
Additionally, consumers have started tracking their own activity on the web using personal health web applications \cite{fernandez2013free} like Microsoft HealthVault or \textit{GoogleFit} as shown in Figure~\ref{fig:workflow}. Some users even provide access to  their Electronic Health Records (EHRs)~\cite{yadav2015mining} for health studies, providing a rich source of information. This data can be utilized by health-care providers as well as consumer electronic companies for risk assessment of subjects, early detection of disorder conditions~\cite{fitcancer}, real-time lifestyle recommendations, and monitoring lifestyle therapy compliance~\cite{yom2017encouraging}.


With the current unprecedented opportunity for understanding the connection between human activity and sleep patterns using wearables technologies, we address following challenges:

\begin{noindlist}
\item \emph{Limited availability of EHR diagnosis data:} Diagnosis information is only available for a few users who allow access to EHRs for health-care providers or application providers. In such a scenario, we have a very limited fraction of users whose both wearables  data  as well as diagnosis information is available. Using a purely supervised learning approach renders the activity data from other users redundant. This necessitates an unsupervised or semi-supervised learning approach that can exploit the larger pool of `unlabeled' activity data coming from wearable devices.  
\item \emph{Exploiting Co-morbidity:} Many disorders are inter-related, with one impacting the other or vice-versa. Caused by common life-style  choices or genetic risks, such mutually co-occurring (and usually correlated) health conditions are referred to as co-morbidity~\cite{berndt1994using}. To exploit this correlational structure, and better understand life-style choices (activity patterns) leading to such outcomes, it is important to jointly learn the models for risk assessment from activity data. 
\item\emph{Generality of learned models:} Due to sample skews of populations in the survey data and available EHRs, along with limited availability,  algorithms developed using only labeled data might not work well with the general population.  In addition, it is also common that general purpose wearable analytics perform poorly in cohorts of people with chronic conditions (\eg\ underestimating steps) \cite{storm2015step}. Hence, an unsupervised learning algorithm that mines the activity patterns rather than doing an end-to-end learning is desirable.
\end{noindlist}

\section{Our Approach} \label{sec:model}

In order to address the problems identified in the previous section, we depict our proposed approach in Figure \ref{fig:workflow}.  Real time activity data collected from consumer wearables through web servers can be used to train our representation learning model, \acttovec. \acttovec\ learns to encode units of activity signals into distributed representations (\aka embeddings) from raw data. These pre-trained embeddings are in turn used to improve performance of the supervised models for the disorder prediction tasks for which diagnosis labels are available only for a small subset. The framework further leverages co-morbidity for multi-task learning on the disorder prediction tasks. In the following, we first describe \acttovec, then we present our supervised model, and finally the multi-task learning setting.




\subsection{Unsupervised Representation Learning} 
\label{subsec:acttovec}
In order to create a representational schema for time-series activity signals, the first natural challenge we encounter is determining the right granularity of the analysis unit. For example, consider the time-series sample in Figure \ref{fig:activity}, where the x-axis represents time at the sampling rate of 30 seconds and the y-axis represents activity levels (or counts), which in our setting are discrete values, ranging roughly from $0$ to $5000$. In continuous values case, a spectrogram like approach can be employed~\cite{greenberg1997modulation}. 

Learning vector representations at the symbol level (\ie\ for each activity level in the y-axis) might result in sparse vectors that are too fine grained  to be effective in the downstream tasks. Similarly, learning a single representation for the entire time sequence (\eg\ spanning a week) could result in generic vectors that lack the required discriminative power to solve the downstream tasks. As we will demonstrate later in our experiments, the right level of granularity is somewhat in between (\eg\ a day span).   

Considering units of analysis shorter than the sequence posits another challenge -- how to capture the contextual dependencies in the representation. Since the units are parts of a sequence that describes a person's activity over a timespan, they are likely to be interdependent. If such dependencies exist, the learning algorithm should capture this in the representation. In the following, we present our representation learning model that addresses these challenges. 


\subsubsection{Granularity of Time-Series Representation} For representing time-series data, it is important to consider the right time-unit for which the embeddings are created. For example, for the activity signal, the granularity of analysis could be at the level of devices' sampling rate (30 seconds in our case), an hour, a day, or a week. Each has its own advantages and disadvantages as mentioned.

\begin{figure}[tp]
  \centering
  \includegraphics[width=0.40\textwidth]{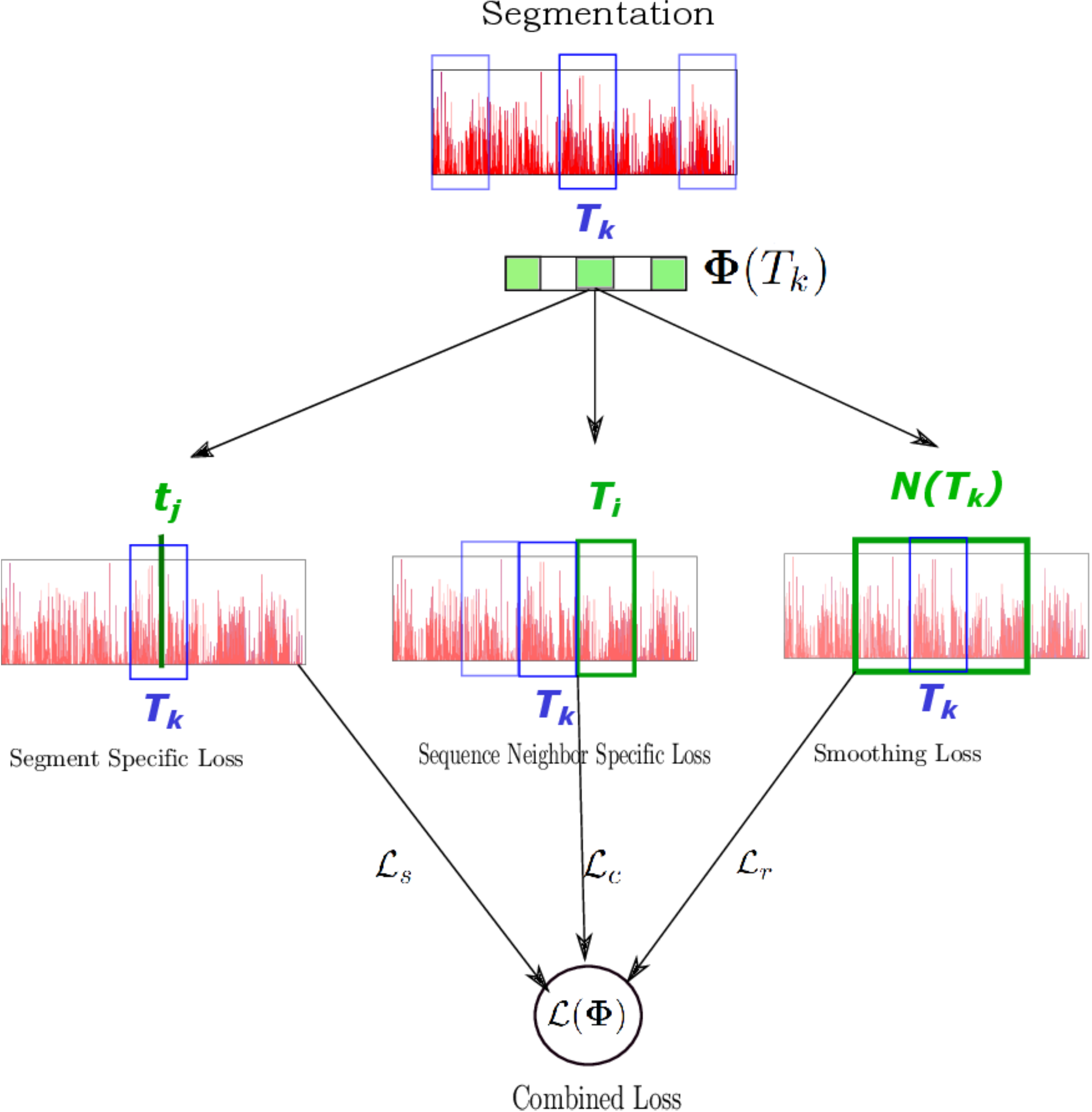}
  \caption{Graphical flow of embedding training by \acttovec's component objectives}
  \label{fig:flowdiag}
\end{figure}

Let $\Ds = \{S_1, S_2, \cdots, S_N\}$ denote a time-series dataset containing activity sequences for $N$ subjects, where each sequence $S_p = (t_1,t_2,\dots,t_n)$  contains $n$ activity measures (\eg\ step counts) for a subject $p$ over a time period (a week in our case). Let $g \in \{$30 seconds, 1 hour, 1 day, 1 week$\}$ specify the granularity of the time span. We first break each sequence $S_p$ into $K$ consecutive time segments of equal length based on the value of $g$ (see at the top of Figure \ref{fig:flowdiag}). Let $T_k = (t_a,t_{a+1},\dots,t_{a+L}) \in \mathcal{T}$ be such a segment of length $L$ that starts at time $a$. Our aim is to learn a mapping function $\ph: \mathcal{T} \rightarrow \real^{d}$ to represent each time segment by a distributed vector representation of $d$ dimensions. Equivalently, the mapping function can be thought of as a look-up operation in an embedding matrix of a single hidden layer neural network (without non-linear activations); and the task is to learn the embedding matrix. The vector representation for a full sequence can then be achieved by concatenating the $K$ segment-level vectors. In this study, we consider the following time spans along with the terminology followed for a comparative analysis:






\begin{noindlist}

\item 30-second samples (\sampletovec): This learns a distributed representation for each 30-second sample given by the device. Hence, our time-series of 20,160 length yields a representational space of $\real^{20160 \times d}$. 




\item Hour (\hourtovec): It learns representation for the chunks of one-hour span of a time sequence, producing a vector space of $\real^{168 \times d}$. 

\item Day (\daytovec): embeds time-series at the level of a day span, giving us a   representational space of $\real^{7 \times d}$.

\item Week (\weektovec): provides embeddings at the scale of a week. A time series of length 20,160, sampled at the rate of 30 seconds, yields a vector in $\real^{d}$ space.  

\end{noindlist}

For a given granularity level, we learn the mapping  function $\ph$ by minimizing a loss that combines three components. Figure~\ref{fig:flowdiag} presents the graphical flow of our model. In the following, we first describe the component losses, and then we present the combined loss. 


%



\subsubsection{Segment-Specific Loss}

We use segment-specific loss to learn a representation for each time segment by predicting its own symbols. This is similar in spirit to the distributed bag-of-words (DBOW) \doctovec\ model of \citeauthor{le2014distributed} (2014), where activity symbols (analogous to `words') and time sequences (analogous to `documents') are assigned unique identifiers, each of which corresponds to a vector (to be learned) in a shared embedding matrix $\ph$. Given an input sequence $T_k = (t_a,t_{a+1},\dots,t_{a+L})$, we first map it to a unique vector $\ph(T_k)$ by looking up the corresponding vector in the shared embedding matrix $\ph$. We then use $\ph(T_k)$ to predict each symbol $t_{j}$ sampled randomly from a window in $T_k$. To compute the prediction loss efficiently, \citeauthor{le2014distributed} use negative-sampling~\cite{mikolov2013distributed}. Formally, the prediction loss with negative sampling is

\begin{equation}\label{eq:granloss}
\begin{split}
\Ls_s(T_k,t_j) &= - \log\sigma(\mathbf{w}_{t_j}^\top \ph (T_k)) \\ 
&\quad - \sum_{m=1}^{M} \mathbb{E}_{t_m \sim \nu(t)} \log\sigma(-\mathbf{w}_{t_m}^\top \ph (T_k))
\end{split}
\end{equation}

\noindent where $\sigma$ is the sigmoid function defined as $\sigma (x) = 1 /(1+ e^{-x})$, $\mathbf{w}_{t_j}$ and $\mathbf{w}_{t_m}$ are the weight vectors associated with ${t_j}$ and ${t_m}$ symbols, respectively, and $\nu(t)$ is the  noise distribution from which ${t_m}$ is sampled. In our experiments, we use unigram distribution raised to the $3/4$ power as our noise distribution, in accordance to \cite{mikolov2013distributed}. 

Since we ask the same segment-level vector to predict the symbols inside the segment, the model captures the overall pattern of a segment. Note that except for \sampletovec, the model learns embeddings for both segments (`sentences') and symbols (`words'). With \sampletovec, in the absence of any higher-level segment, the model boils down to the Skip-gram \wordtovec\ model \cite{mikolov2013distributed} that learns embeddings for the symbols using a window-based approach. It is important to mention that segment-based approach is commonly used in time-series analysis, though among the representational models only vector space models like SAX-VSM \cite{senin2013sax} look at the co-occurrence statistics at the segment level (indirectly), with a \textit{bag-of-words} assumption.

\subsubsection{Sequence-Neighbor Specific Loss}

The previous objective in Equation~\ref{eq:granloss} captures local patterns in a segment. However, since the segments are contiguous and describe activities of the same person, they are likely to be related. For example, after a strenuous hour or day, there might be lighter activity periods. Therefore, representation learning algorithms should capture such relations between nearby segments in a time-series. We formulate this relation by asking the current segment vector $\ph(T_k)$ (to be estimated) to predict its neighboring segments in the time-series: $\ph(T_{k-1})$ and $\ph(T_{k+1})$. Recall that each segment is assigned a unique identifier. If $T_i$ is a neighbor to $T_k$, the neighbor prediction loss using negative sampling can be formally written as: 


\begin{equation}\label{eq:conloss}
\begin{split}
\Ls_c(T_k,T_i) &= - \log\sigma(\mathbf{w}_{T_i}^\top \ph (T_k)) \\ 
&\quad - \sum_{m=1}^{M} \mathbb{E}_{T_m\sim \nu(T)} \log\sigma(-\mathbf{w}_{T_m}^ \ph (T_k))
\end{split}
\end{equation}

\noindent where, $\mathbf{w}_{T_i}$ and $\mathbf{w}_{T_m}$ are the weight vectors associated with $T_i$ and $T_m$ segments in the embedding matrix, respectively, and $\nu(T)$ is the unigram noise distribution over sequence IDs. As before, the noise distribution $P(T)$ for negative sampling is defined as a unigram distribution of sequences raised to the $3/4$ power. 


\subsubsection{Smoothing Loss}

While the previous two objectives attempt to capture local and global patterns in a time series, we also hypothesize that there is a smoothness pattern between neighboring segments.  In some sense, it can also be viewed as a way to capture the periodicity of human activity. The learning algorithm should discourage any abrupt changes in the representation of nearby segments. We formulate this by minimizing the $l_2$-distance between the vectors. Formally, the smoothing loss for a time-segment $T_k$ is

\begin{equation}\label{eq:regloss}
\begin{split}
\Ls_r(T_k,\Ns(T_k)) &= \frac{\eta}{\mid \Ns(T_k) \mid} \sum_{T_c \in \Ns(T_k)} \norm{\ph(T_k) - \ph(T_c)}^2 
\end{split}
\end{equation}

\noindent where, $\Ns(T_k)$ is the set of time-segments in proximity to $T_k$ and $\eta$ is the smoothing strength parameter. Note that the smoothing loss is not applicable to \weektovec.





\subsubsection{Combined Loss}

We define our \acttovec\ model as the combination of the losses described in Equations~\ref{eq:granloss}, ~\ref{eq:conloss}, and ~\ref{eq:regloss}:


\setlength{\textfloatsep}{0.15in}
\begin{algorithm}[t!]
\footnotesize
\SetKwInOut{Input}{Input}\SetKwInOut{Output}{Output}
\SetAlgoNoLine
\SetNlSkip{0em}
\Input{set of time-series  $\Ds = \{S_1, S_2, \cdots, S_N\}$ with $S_p=(t_1,t_2,\cdots,t_n)$, granularity level $g$}
\Output{learned time-series representation $\mathbf{\Phi}(S_p)$}
1. Break each time-series $S_p$ into segments based on the granularity $g$;\\
2. Initialize parameters: $\mathbf{\Phi}$ and $\mathbf{w}$'s; \\
3. Compute noise distributions: $\nu(t)$ and $\nu(T)$ \\ 
4. \Repeat {convergence}{
- Permute $\Ds$;\\
\For {each time-series sequence $S_p \in \Ds$}{
\For {each time-segment $T_k \in S_p$}{ 
    \hspace{-0.0cm}\For{each time-series sample $t_j \in T_k$}{ 
  		\hspace{-0.3cm}- Consider $(T_k, t_j)$ as a  positive pair and generate $M$ negative pairs $\{({T_k}, t_m)\}_{m=1}^{M}$ by sampling $t_m$ from $\nu(t)$; \\
  		\hspace{-0.3cm}- Perform gradient update for $\Ls_s (T_k, t_j)$; \\
  		\hspace{-0.3cm}- Sample a neighboring time-segment $T_i$ from sequence $S_p$; \\
  		\hspace{-0.3cm}- Consider $(T_k, T_i)$ as a  positive pair and generate $M$ negative pairs $\{({T_k}, T_m)\}_{m=1}^{M}$ by sampling $T_m$ from $\nu(T)$; 
        
  		\hspace{-0.3cm}- Perform gradient update for $\Ls_c (T_k, T_i)$; \\
    	\hspace{-0.3cm}- Perform gradient update for $\Ls_r (T_k, \Ns(T_k))$;       
    } 
    }
   }
}   
\caption{Training \acttovec\ with SGD}
\label{alg:jnt}
\end{algorithm}

\begin{equation}\label{eq:totalloss}
\begin{split}
\Ls(\ph) &= \sum_{p=1}^P \sum_{\substack{T_k \in S_p}} \sum_{\substack{t_j \in T_k\\ T_i \in \Ns(T_k)}} \Big[\Ls_s(T_k,t_j) +\\ &\quad \Ls_c(T_k,T_i)  + \Ls_r(T_k, \Ns(T_k))\Big]
\end{split}
\end{equation}

We train the model using stochastic gradient descent (SGD); Algorithm \ref{alg:jnt} gives a pseudocode. We first initialize the model parameters $\ph$ and $\mathbf{w}$ with small random numbers sampled from uniform distribution $\Us(-0.5/d,  0.5/d)$, and compute the noise distributions $\nu(t)$ and $\nu(T)$ for $\Ls_s(T_k,t_j)$  and $\Ls_c(T_k,T_i)$ losses, respectively.

To estimate the representation of a segment, for each symbol sampled randomly from the segment, we take three successive gradient steps to account for the three loss components in Equation \ref{eq:totalloss}. By making the same number of gradient updates, the algorithm weights equally the contributions from the symbols in a segment and from the neighbors. Note that for \sampletovec~and \weektovec~ only $\Ls_c$ loss is calculated since the other two objectives do not apply. 

\begin{figure*}[t!]
  \centering
  \includegraphics[width=0.8\textwidth]{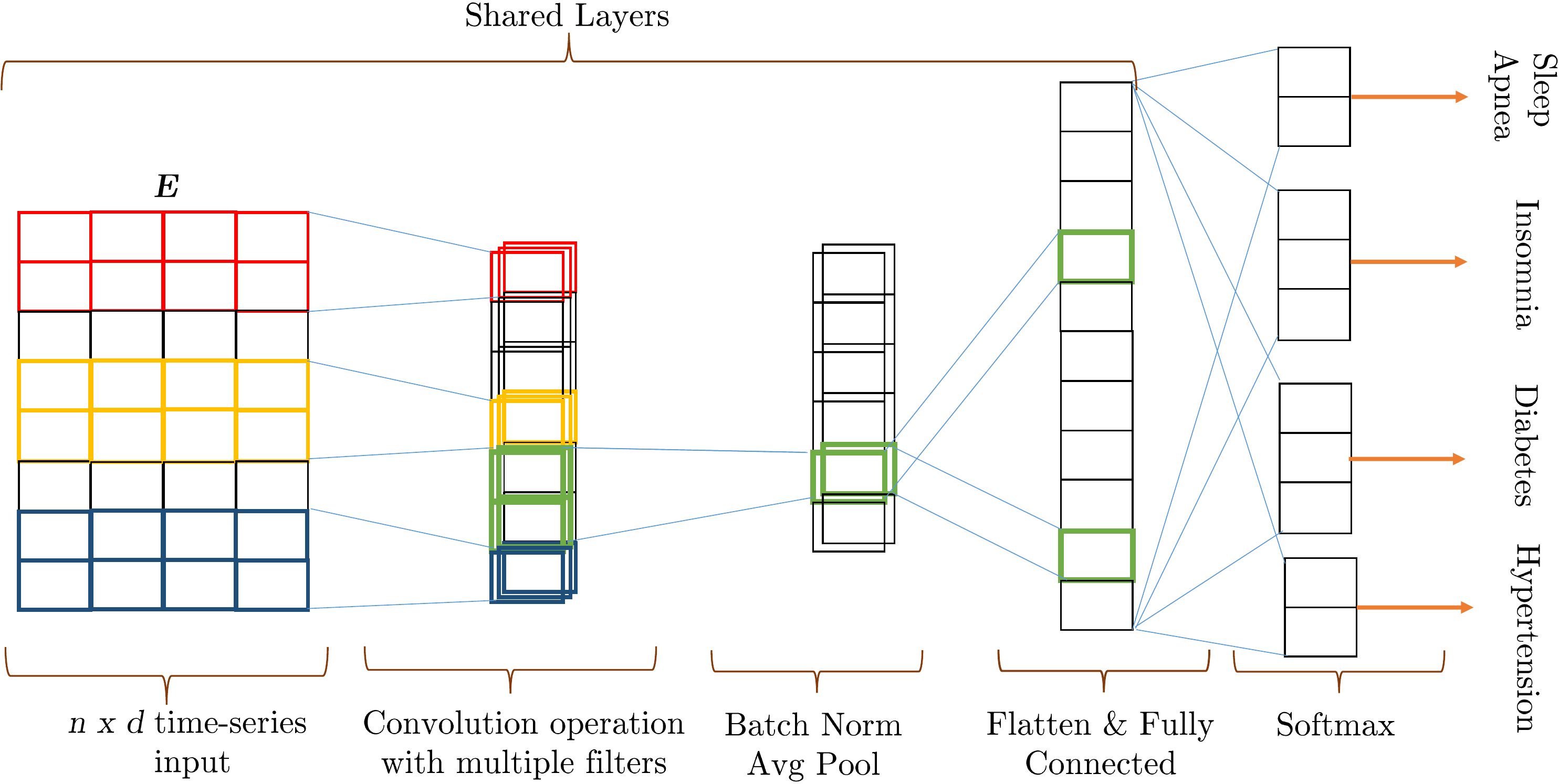}
  \caption{Multi-task learning deep convolution neural network with batch normalization and average pooling operations.}
  \label{fig:multicnn}
\end{figure*}

\subsection{Supervised Multi-task Learning} \label{subsec:cnn}

In recent years, deep neural networks (DNNs) have shown impressive performance gains in a wide spectrum of machine learning problems such as image recognition, language translation, speech recognition, natural language parsing, bioinformatics, and so on. Apart from the improved performance, one crucial benefit of DNNs is that they obviate the need for feature engineering and learn latent task-specific features automatically as distributed dense vectors. Recently, DNNs have also been successfully applied to classification problems with time-series data~\cite{lipton2015learning,wang16,razavian2016multi,sathyanarayana2016impact, zheng2016exploiting}.

\subsubsection{Disorder Prediction with Convolutional Neural Network} 

In our work, we use a convolutional neural network as it has shown impressive results on similar tasks with time-series data \cite{wang16,sathyanarayana2016impact}. Figure \ref{fig:multicnn} shows our network. The input to the network is a time sequence $S_p=(t_1,t_2,\cdots,t_n)$ containing activity symbols coming from a finite vocabulary $\Vs$. The first layer of our network maps each of these symbols into a distributed representation in $\real^d$ by looking up a shared embedding matrix $E$ $\in$ $\real^{|\Vs| \times d}$.  We can initialize $E$ randomly or using pre-trained \acttovec\ vectors. The output of the look-up layer is a matrix  $X \in \real^{n \times d}$, which is passed through a number of convolution and pooling layers to learn higher-level feature representations. A \emph{convolution} operation applies a \emph{filter} $\mathbf{u} \in \real^{k.d}$ to a window of $k$ vectors to produce a new feature, $h_i = f(\mathbf{u} . X_{i:i+k-1})$, where $X_{i:i+k-1}$ is the concatenation of $k$ look-up vectors, and $f$ is a nonlinear activation; we use rectified linear units or ReLU. We apply this filter to each possible $k$-length windows in $X$ with stride size of $1$ to generate a \emph{feature map}, $\mathbf{h}^j = [h_1, \ldots, h_{n+k-1}]$. 

We repeat the above process $N$ times with $N$ different filters to get $N$ different feature maps. We use a \emph{wide} convolution \cite{Kalchbrenner14}, which ensures that the filters reach the entire sequence, including the boundary symbols. This is done by performing \emph{zero-padding}, where out-of-range (\ie\ $i<1$ or $i>n$) vectors are assumed to be zero. With wide convolution, $o$ zero-padding size and $1$ stride size, each feature map contains $(n + 2o - k + 1)$ convoluted features. After the convolution, we apply an \emph{average-pooling} operation to each of the feature maps to get $\mathbf{m} = [\mu_l(\mathbf{h}^1), \cdots, \mu_l(\mathbf{h}^N)]$, where $\mu_l(\mathbf{h}^j)$ refers to the average operation applied to each window of $l$ features with stride size of $1$ in the feature map $\mathbf{h}^j$. 
Intuitively, the convolution operation composes local features into higher-level representations in the feature maps, and average-pooling extracts the important aspects of each feature map while reducing the output dimensionality. Since each convolution-pooling operation is performed independently, the features extracted become invariant in order (\ie\ where they occur in the time sequence). To incorporate order information between the pooled features, we include a fully-connected layer $\mathbf{z} = f(V\mathbf{m})$ with $V$ being the weight matrix. Finally, the output layer performs the classification. Formally, the classification layer defines a Softmax
\vspace{-0.4em}
\begin{eqnarray}
p(y=k|S_p, \theta) = \frac{\exp (W_{k}^T \mathbf{z} )} {\sum_{k'} \exp ( {W}_{k'}^T \mathbf{z} )} \label{eq:softmax}
\end{eqnarray}

\noindent where ${W}_{k}$ are the class weights, and $\theta = \{E, U, V, W\}$ defines the model parameters. We use a cross-entropy loss 
\vspace{-0.4em}
\begin{eqnarray}
\Ls(\theta) = {- \frac{1}{B} \sum_{i=1}^{B} \mathbb{I} (y_i=k) \log \hat{y}_{ik}} + \frac{\lambda_2}{2} \norm{\mathbf{W}}^2 + \lambda_1 \norm{\mathbf{W}}_1 \label{eq:celoss}
\end{eqnarray}


\noindent where $\hat{y}_{ik} = p(y_i=k|S_p, \theta)$, $B$ is the batch size, and $\mathbb{I}(.)$ is an indicator function that returns $1$ when the argument is true, otherwise it returns $0$. We use elastic-net regularization on the last layer weights $W$, with $\lambda_1$ and $\lambda_2$ being the strengths for the $L_1$ and $L_2$ regularizations, respectively. Additionally, we use \emph{batch-normalization}~\cite{ioffe2015batch} and \emph{dropout}~\cite{srivastava2014dropout} to regularize the network. Both these methods have shown to work well in practice reducing over-fitting, with batch normalization providing faster convergence.



\subsubsection{Multi-Task Learning with Shared Layers} \label{subsec:multitask}

We exploit co-morbidity through multi-task learning with the assumption that joint training for multiple related tasks can improve the classification performance by reducing the generalization error. This idea has been successfully employed in natural language processing~\cite{collobert2011natural}, speech recognition~\cite{deng2013new}, and for 
clinical visits~\cite{lipton2015learning,razavian2016multi}.

Our approach is similar in spirit to the approach of \cite{collobert2011natural}, where the models for different tasks share their parameters. As shown in Figure \ref{fig:multicnn}, the models for the four different disorder prediction tasks share their embedding, convolution-pooling, and fully-connected layers, comprising parameters $\theta_s = \{E, U, V\}$, and each task has its own weight matrices $W$ for the Softmax outputs. Formally, the prediction for task $m$ can be written as $p(y_m = k | S_p, \theta_s, W_m)$ (see Equation \ref{eq:softmax}), where $y_m$ and $W_m$ denote the output variable and the Softmax weight matrix, respectively, associated with task $m$. The overall loss of the combined model can be written as
\vspace{-0.4em}

\begin{equation}\label{eq:multiloss}
\begin{split}
\Ls_m (\theta) &=  \sum_{m=1}^{M} \alpha_m \Ls_m (\theta_s, W_m)  
\end{split}
\end{equation}
\vspace{-0.4em}

\noindent where $\Ls_m (\theta_s, W_m)$ is the loss for task $m$ (see Equation \ref{eq:celoss}), and $\alpha_m$ is its mixture weight. In our case, $M$=4, over the four disorder prediction tasks: sleep apnea, insomnia, diabetes, and hypertension. Multi-task Learning has been shown to increase performance on individual tasks by utilizing additional information from the auxiliary tasks, making the model more generalizable.  

\textit{Remark:} We made an additional attempt to jointly model the output random variables with a global inference (decoding) inside the learning algorithm. This is in fact equivalent to putting a fully-connected Conditional Random Field or CRF (with node and edge potentials, and a global normalization term) in the output layer of the network. While such methods improve results in general (\eg\ \cite{zheng2015conditional,Sutton:2005b}), in our problem, we observed that CRF induced DNN does not produce performance gains. Similar negative results were reported by \cite{Sutton:2005} for multi-task learning with joint decoding.   





\section{Experimental Settings} \label{sec:settings}


In this section, we describe our experimental settings -- the prediction tasks on which we  evaluate the learned embeddings, the datasets, the models we compare, and their settings. 
\subsection{Human activity time-series}
The human activity data collected with a wearable device (actigraphy) records mean activity count per base time-unit depending on the sampling rate of the device. The datasets we are working on, as described in next section, provide us with a signal that can only take integer values. Unlike most time-series data, this makes embedding the input straightforward, without any pre-processing for our proposed \acttovec\ method. In case of floating point value signals, a preprocessing step of waveform extraction can be added, as done by speech recognition community~\cite{greenberg1997modulation}. Actigraphy data is widely used for diagnosis of sleep disorders and quantification of physical activity for epidemiological studies.

\subsection{Datasets}

We use the Study of Latinos (SOL)~\cite{sorlie2010design} and 
the Multi-Ethnic Study of Atherosclerosis (MESA)~\cite{bild2002multi} 
datasets. These datasets are made publicly available as a part of initiative 
to provide computer scientists with resources that can be used for helping the 
clinical experts~\cite{dean2016scaling}. SOL has data for 1887 subjects 
ranging from physical activity to general diagnostic tests, while MESA has 
mostly the activity time-series data for 2237 subjects. Hence, this simulates 
the scenario described in Section 3 and Figure~\ref{fig:workflow} with diagnosis 
labels available only for a proportion of wearables consumers. National Sleep 
Research Resource provides these datasets at \url{sleepdata.org} with 
activity data (actigraphy) per subject for a minimum of 7 days measured with 
wrist-worn Philip's Actiwatch Spectrum device. Essentially, we get 28,868 days of actigraphy data across subjects to train our model on. Both the datasets contain 
time-series of activity counts for each subject sampled at a rate of 30 
seconds. Figure~\ref{fig:activity} presents a sample activity signal.

A total of 1757 
discrete values of signal were observed for our combined dataset. 
A very few missing values were observed in the dataset; those were 
replaced by unknown (UNK) token while training our model. We 
only considered 7 days of data for each subject, since missing 
values increased enormously for subjects with more than 7 days of 
data. Any unseen or out-of-vocabulary signal value can be handled by a 
procedure like assigning representation from averaging out 
neighboring signal values rather than a generic unknown symbol 
assignment.

\emph{Note}: all the diagnosis prediction tasks were taken only from the SOL 
dataset, since MESA does not have the diagnosis data, public. We just use 
actigrpahy time-series from MESA for creating our \acttovec\ embeddings.

\subsection{Prediction Tasks}

We evaluate the effectiveness of the learned embeddings on the following physical and mental disorder prediction tasks:  

\begin{noindlist}
\item \textbf{Sleep Apnea:} Sleep apnea syndrome is a sleep disorder characterized by reduced respiration during the sleep time, reducing oxygen flow to body. We use the Apnea Hypopnea Index (AHI) at 3\% desaturation level with AHI$<$5 being characterized as \emph{non-apneaic}, while  AHI $>5$ indicating a \emph{mild-to-severe-apnea}. 
\item \textbf{Diabetes:} Diabetes (type 2) is inability of body to respond to insulin, leading to elevated levels of blood sugar. Diabetes prediction is defined as a three-class classification problem, where the task is to decide whether a subject is a \emph{non-diabetic}, \emph{pre-diabetic}, or \emph{diabetic}. 

\item \textbf{Hypertension:} Hypertension refers to abnormally high levels of blood pressure, an indicator of stress. Hypertension prediction characterizes a binary classification problem for increased blood pressure (BP). BP $>$ 140/90 is considered as having \emph{hypertension}.

\item \textbf{Insomnia:} Insomnia is a sleep disorder characterized by inability to fall sleep easily, leading to low energy levels during the day. We use a 3-class prediction problem for classifying subjects into \emph{non-insomniac}, \emph{pre-insomniac} and \emph{insomniac} groups. We merged subjects suffering from moderate and severe insomnia into one class owing to very few subjects suffering from severe insomnia. 
\end{noindlist}

\subsection{Models Compared}

We compare our method with a number of naive baselines and existing systems that use symbolic representations:

\subsubsection{Baselines}

\paragraph{\Ni Majority Class} This baseline always predicts the class that is most frequent in a dataset. 

\paragraph{\Nii Random} This baseline randomly picks a class label.

\paragraph{\Niii SAX VSM:} Symbolic Aggregate Representation Vector Space or SAX-VSM ~\cite{senin2013sax} combines SAX~\cite{lin2007experiencing}, one of the most  widely used symbolic representation technique for time-series data with Vector Space Modeling using tf-idf (term frequency inverse document frequency) measure. 

\paragraph{\Niv BOSS:} Bag-of-Symbolic-Fourier-Approximation or BOSS~\cite{schafer2015boss} 
is a symbolic  representational learning technique that uses Discrete Fourier 
Transform (DFT) on sliding windows of time-series. 
BOSS creates histograms of 
Fourier coefficients to create equal sized bins of the Fourier coefficients 
over the time-series, which are then assigned representational symbols. The 
classification method involves nearest neighbor approach, with labels assigned 
based on class that gets highest similarity score.  

\paragraph{\Nv BOSSVS:} BOSS in Vector Space or 
BOSSVS~\cite{schafer2016scalable} is a vector space model similar to SAX-VSM 
except that it uses tf-idf vector space of the symbolic representation of the 
time-series obtained through BOSS. BOSS is known to be one of the most accurate  method 
on standard time-series classification tasks, with BOSS-VS performing 
marginally lower.  

\subsubsection{Variants of \acttovec}

We experiment with the following variants of our unsupervised learning model:

\paragraph{\Ni Unregularized models:} This group of models omit the smoothing component $\Ls_r(T_k,\Ns(T_k))$ in Equation \ref{eq:totalloss}. In the Results section, we refer to these models as \sampletovec, \hourtovec, \daytovec, and \weektovec. 


\paragraph{\Nii Regularized models:} We perform smoothing in these models. This group includes \hourtovecreg~and \daytovecreg. We omit \sampletovecreg~since it performed extremely poorly on all the tasks. Recall that smoothing is not applicable to \weektovec.


\subsubsection{Supervised Learning Variants}
\paragraph{\Ni Task-specific models:} As described earlier, these models are trained end-to-end for the disorder task at hand. 
\paragraph{\Nii Multi-task Learning models:} These models are trained jointly with all the disorder prediction tasks learnt jointly.
\paragraph{\Niii Pre-trained models:} These models are initialized with  embeddings from best performing  \daytovecreg\ . 

\subsection{Hyper-parameter selection}

For hyper-parameter tuning, we use development set containing 10\% of the data for all the experiments. We have the following hyper-parameters for \acttovec: window size ($w$) for segment-specific  loss, number of neighboring segments ($|\Ns(T_k)|$) and regularization strength ($\eta$) for \daytovec\ and \hourtovec. We tuned for $w \in \{8,12,20,30,50,120, 500\}$, $\eta \in \{0,0.25,0.5,0.75,1\}$, and $|\Ns(T_k)| \in \{2,4\}$. We chose $w$ of size 20, 20, 30, and 50 for \sampletovec, \hourtovec, \daytovec, and \weektovec, respectively. The $\eta$ of 0.25 and 0.5 were chosen for \daytovec\ and \hourtovec, respectively. The neighbor set size of 2 was selected for all the models. We selected an embedding size $d$=$100$ for all our models. 

Dropout rate of 0.5 was selected for all the supervised tasks with CNN. We used 
Adam Optimizer for all our supervised learning tasks. We tuned $\lambda_1, \lambda_2 \in \{0,0.25,0.5,1\}$ for our tasks. We optimize multi-task weights, 
$\alpha$, such that sum of weights is always one. For the multi-task learning 
without initialization, we used $\alpha = \{0.2,0.2,0.4,0.2\}$, respectively, while 
with pre-trained embeddings with multi-task learning, 
we settled for $\alpha = \{0.3,0.25,0.35,0.15\}$, respectively for sleep-apnea, diabetes, insomnia, and hypertension.
We used a 3, 4,3,3, and 3 layered CNN for sleep-apnea, diabetes, insomnia,
hypertension, and multi-task learning tasks, respectively. In the next section, we describe our findings on the test dataset.

\begin{table}[tp]
  \renewcommand{\arraystretch}{0.9}
  \caption{\textbf{Acc}uracy, \textbf{Pre}cision, \textbf{Rec}all, \textbf{Spec}ificity, and $\mathbf{F_1}$ values for \textbf{Sleep-Apnea} prediction for each method. \texttt{+Pre} indicates pre-trained embeddings.}
  \label{ApneaResults}
  \centering
  \vspace*{-\baselineskip}
  \resizebox{\columnwidth}{!}{%
  \begin{tabular}
{l|l|ccccc}
	\toprule
    Method & Clf. & Acc. & Pre. & Rec. & Spec. & $F_{1}$ \\
    \midrule
    \texttt{Majority} & 0-R & 74.6 & 00.0 & 00.0 & 100.0 & 00.0\\
    \texttt{Random} & & 50.0 & 25.6 & 50.0 & 50.0 & 33.9 \\
    
    \midrule
    \texttt{SAX-VSM} & & 74.6 & 00.0 & 00.0 & \emph{100.0} & 00.0 \\   
    \texttt{BOSS} & &  70.4 & 30.0  & 12.5 & 90.1 & 17.6\\  
    \texttt{BOSSVS} & & 68.2& 20.0 & 8.3  & 88.6  & 11.7\\  
    \midrule
	\sampletovec & LR & 50.0 & 27.8 &	54.0 & 48.5 & 36.7\\ 
    \hourtovec & LR  & 70.3  & 46.1 &	22.2 &  89.6 & 30.0\\ 
    \hourtovecreg & LR & 71.4  & 36.8  &	14.3 & 91.4 & 20.5\\ 
    \daytovec & {LR} & 61.9 & 32.8  &	42.0 & 69.1 & 36.8\\
    \daytovecreg & {LR} & 65.1  & 39.6 & 38.2 & 76.1 & 38.9\\
    \weektovec & {LR} & \emph{75.1} & \emph{57.1} & 8.3 & 97.9 & 14.5  \\ \midrule
    \texttt{Task-spec} & {CNN} & 55.3 & 31.0 & 62.8 & 52.7 & 41.5\\ 
    \texttt{Task-spec+Pre} & {CNN} & 68.2 & 39.6 & 47.5 & 75.4 & 43.2\\
     \texttt{Multi-task} & {CNN} & 54.7 & 31.9 & \emph{69.8} & 49.6 & 43.8\\
      \texttt{Multi-task+Pre} & {CNN} & 65.9 & 37.7 & 53.5 & 70.1 & \textbf{44.2}\\\bottomrule
\end{tabular}
}
\end{table}

\section{Results} \label{sec:results}
In this section, we present our results for the four prediction tasks. The results are presented in Tables~\ref{ApneaResults},~\ref{diabResults},~\ref{insomniaResults}, and~\ref{hyperResults} in four groups: \Ni baselines, \Nii existing symbolic methods, \Niii our \acttovec\ variants, and \Niv our supervised variants. We first discuss the results obtained with unsupervised representational learning models.

\subsection{Unsupervised Representation Learning}

Since our goal is to evaluate the effectiveness of the learned vectors, we use simple linear classifiers to predict the class labels. Primarily, we use a Logistic Regression (LR) classifier with our \acttovec\ models. For the multi-class classification problems like Diabetes and Insomnia, we use One-vs-All classifiers, tuning for micro-$F_{1}$ score. We ran each experiment 10 times and take the average of the evaluation measures to avoid any randomness in results. 



\begin{table}[tp]
  \renewcommand{\arraystretch}{0.9}
  \caption{  \textbf{Pre}cision (weighted), \textbf{Rec}all (weighted), $F_1$ values (weighted), and $F_1$-micro scores for the three class classification --- non-diabetic, pre-diabetic, and diabetic --- of \textbf{diabetes} prediction for each methods. \texttt{+Pre} indicates pre-trained embeddings. }
  \label{diabResults}
  \centering
    \vspace*{-\baselineskip}
  \resizebox{\columnwidth}{!}{%
  \begin{tabular}{l|l|cccc}
    \midrule

    Method & Clf. & Pre. & Rec. & $F_{1}$-macro & $F_{1}$-micro \\
    
    \midrule
    \texttt{Majority} & 0-R & 23.7 & 48.7 & 21.7 & 31.9 \\
    \texttt{Random} & & 37.7 & 33.3 & 33.3 & 33.3  \\
    
    \midrule
    \texttt{SAX-VSM} & & 34.4 & 43.9 & 38.6 & 24.3  \\   
    \texttt{BOSS} & & 39.1 & 38.8 & 38.9  & 31.5 \\  
    \texttt{BOSSVS} & & 39.6 & 40.7 & 40.1 & 32.7 \\  
        \midrule
    \sampletovec & LR &  41.2  & 38.9 & 40.0 & 36.7 \\ 
    \hourtovec & LR & 39.5 & 44.4 &	41.4 & 33.3\\ 
    \hourtovecreg & LR & 40.8 &    43.9 &  42.1 & 32.0 \\ 
    \daytovec & LR & 41.2 & 40.7 &	40.9 & 38.0\\ 
    \daytovecreg & LR  & 44.7 & 40.7 &	41.8  & \textbf{39.5}\\ 
    \weektovec & {LR} &  40.8 &    44.4   &  40.6 & 34.1\\
    \midrule
\texttt{Task-spec} & {CNN} & 40.0 & \emph{51.4} & 45.2 & 41.0\\ 
    \texttt{Task-spec+Pre} & {CNN} & 45.8 & 46.4 & 44.6 & 41.7 \\
     \texttt{Multi-task} & {CNN} & 46.1 & 47.1 & 45.6 & 43.7\\
      \texttt{Multi-task+Pre} & {CNN} & \emph{46.8} & 47.8 & \emph{46.5} & \textbf{44.4}\\\bottomrule
      \end{tabular}
      }
\end{table}

As can be observed, across all the tasks, the \daytovecreg~outperforms 
all the models including the baseline time-series models. Across the board, models 
involving granularity on the scale of a day performs better than all the other 
granularities as well as baseline time-series methods. Clearly, among the \acttovec\ variants, the \weektovec~models perform the worst, while \hourtovec\ models 
perform just a bit better on an average. Hour- and week-level models 
perform around the same as the baseline time-series methods. The high-dimensional 
models based on samples (\ie\ \sampletovec) perform better than hour-level, week-level, 
and baseline models. \daytovec\ produces marginally better results than the \sampletovec\ despite much lower dimensional space (2880x). 

Intuition behind adding the smoothing loss to our model 
with Equation~\ref{eq:regloss} was to test the hypothesis that 
periodicity in human activities should be reflected in neighboring time-segments, which should be similar in structure representing a continuity. 
As can be observed from the results, the regularization hypothesis was 
misguided at the sample- and hour-level segments. 
However, adding regularization helps produce gains across the board at the level of \daytovec\ , our best \acttovec\ model. 
 
 \daytovec\ consistently gives 2-4\% (absolute) better than our other \acttovec\ models, 6-10\% (absolute) on best of majority/random, and 6-20\% (absolute) than the baseline time-series models on $F_1$ scores on all tasks.

Another important aspect to note is the increase in generalization across classes on the prediction task of \hourtovec\ and \hourtovecreg\ . Our datasets are 
imbalanced with majority class being the subjects not suffering from the disorders
under consideration, with classification task being to predict the disorder-positive
subjects. Most of our models and baselines are highly biased towards predicting
the majority class. \hourtovec~has lower accuracy but higher precision and recall 
than most of models, owing to its lower bias. Regularized \hourtovecreg~does 
better on $F_1$ scores while increasing the specificity/micro-$F_1$ scores along 
with accuracy on all the prediction tasks than \hourtovec, thus making it more 
generalized.

Clearly, the level of granularity makes a lot of difference to the performance of 
our models. From the above results on four different tasks we can conclude that 
while low granularity level (\eg \sampletovec) suffered from coarse
embeddings, the high granularity 
(\weektovec) level embeddings lost the ability to discriminate. 

\begin{table}[tp]
\vspace{-1em}

  \renewcommand{\arraystretch}{0.9}
  \caption{  \textbf{Pre}cision (weighted), \textbf{Rec}all (weighted), $F_1$ (weighted), and $F_1$-micro scores for three class classification --- no-insomnia, pre-insomnia, and (moderate+severe) insomnia --- of \textbf{insomnia} prediction. \texttt{+Pre} indicates pre-trained embeddings.}
  \label{insomniaResults}
  \centering
    \vspace*{-\baselineskip}
  \resizebox{\columnwidth}{!}{%
  \begin{tabular}{l|l|cccc}
    \midrule

    Method & Clf. & Pre. & Rec. & $F_{1}$-macro & $F_{1}$-micro \\
    \midrule
    \texttt{Majority} & 0-R & 38.3 & 61.9 & 47.4 & 25.5\\
    \texttt{Random} & & 46.6 & 33.3 & 33.3 & 33.3  \\
    \midrule
    \texttt{SAX-VSM} & & 38.3 & 61.9 & 47.4 & 25.5  \\   
    \texttt{BOSS} & & 47.6 & 52.2 & 49.8 & 34.9 \\  
    \texttt{BOSSVS} & & 45.2 & 50.1& 47.5  & 33.1 \\  \midrule
    \sampletovec & LR &  41.6   &  43.9   &  42.4 & 35.3\\ 
    \hourtovec & LR & 42.5 & 52.4 &	44.6 & 28.5\\ 
    \hourtovecreg & LR &  39.8   & 51.3  &  43.5 & 28.7 \\ 
    \daytovec & LR & 46.2 & 44.4 &	45.2 & 35.8 \\ 
    \daytovecreg & LR  & 47.9 & 45.5 &	46.6 & \textbf{39.7}\\ 
    \weektovec & {LR} & 51.5   &  55.0  &   44.2 & 31.5\\\midrule
\texttt{Task-spec} & {CNN} & 50.9 & 50.6 & 50.7 & 40.1\\ 
    \texttt{Task-spec+Pre} & {CNN} & 54.5 & 58.2 & 55.6 & 41.2 \\
     \texttt{Multi-task} & {CNN} & 55.7 & \emph{66.5} & 56.3 & 41.2\\
      \texttt{Multi-task+Pre} & {CNN} & \emph{58.3} & 65.8 & \emph{56.5} & \textbf{41.7}\\\bottomrule

      \end{tabular}
      }
\end{table}

\begin{table}[tp]
\vspace{-1em}
  \renewcommand{\arraystretch}{0.9}
  \caption{\textbf{Acc}uracy, \textbf{Pre}cision, \textbf{Rec}all, \textbf{Spec}ificity, and $\mathbf{F_1}$ values for \textbf{Hypertension} prediction for each method. \texttt{+Pre} indicates pre-trained embeddings.}
  \label{hyperResults}
  \centering
    \vspace*{-\baselineskip}
  \resizebox{\columnwidth}{!}{%
  \begin{tabular}
{l|l|ccccc}
	\toprule
    Method & Clf. & Acc. & Pre. & Rec. & Spec. & $F_{1}$ \\
    \midrule
    \texttt{Majority} & 0-R & 74.9 & 00.0 & 00.0 & 100.0 & 00.0\\
    \texttt{Random} & & 50.0 & 25.1 & 50.0 & 50.0 & 33.4 \\
    
    \midrule
    \texttt{SAX-VSM} & & 74.9 & 0.00 & 0.00 & \emph{100.0} & 0.00 \\   
    \texttt{BOSS} &  & \emph{69.9} & 35.2 & 25.5 & 84.5 & 29.6\\  
    \texttt{BOSSVS} & & 69.9 & 36.1 & 27.7 & 83.8 & 31.3\\  \midrule
    \sampletovec & LR & 51.3 & 33.3  & 48.3 & 52.7 & 39.5\\ 
    \hourtovec & LR  & 68.2 & 36.7 & 18.4 & 87.4 & 24.4 \\ 
    \hourtovecreg & LR  & 68.2 & 36.0 & 17.0 & 88.2 & 23.1 \\ 
    \daytovec & LR  & 60.8 & 39.1 &	41.7 & 69.8 & 40.3\\
    \daytovecreg & LR  & 68.2 & 41.8 &	45.0 & 76.8 & \textbf{43.4}\\
    \weektovec & {LR} & 67.7 & \emph{58.3} & 11.1 & 96.0 & 18.7 \\ 
    \midrule
    \texttt{Task-spec} & {CNN} & 69.4 & 44.1 & 31.2 & 84.4 & 36.6\\ 
    \texttt{Task-spec+Pre} & {CNN} & 65.8 & 39.1 & 37.5 & 77.0 & 38.3\\
     \texttt{Multi-task} & {CNN} & 61.1 & 47.5 & 40.6 & 82.7 & 43.8\\
      \texttt{Multi-task+Pre} & {CNN} & 61.7 & 38.0 & \emph{56.2} & 45.3 & \textbf{44.2}\\\bottomrule
      \end{tabular}
      }
\end{table}

%
    

\subsection{Supervised Learning}
Results obtained on supervised learning models are shown in 
Tables~\ref{ApneaResults},~\ref{diabResults},~\ref{insomniaResults}, and~\ref{hyperResults}; see the last group of results. Barring the exception of hypertension, all the task-specific end-to-end convolution neural networks (CNNs) perform better than \acttovec~'s logistic classifier. This is not surprising since the CNNs are directly trained on the task. Using pre-trained embeddings from the \acttovec\ boosts the  $F1$ 
scores of task-specific CNNs across the board by ~1\%-2\% (absolute). 

Using joint multi-task learning improves the performance 
for the downstream tasks by 1\%-5\% (absolute) in $F1$ scores, notable 
being hypertension task, compared to its counterpart with task-specific
classification. As observed with task-specific learning, using pre-trained embeddings improves the performance of multi-task learning further by ~1\%-2\% (absolute).

Using pre-trained embeddings improves the 
performance of our methods as have been observed in a number of other 
domains. This is especially significant for our problem, since the 
disorder diagnosis data is available only for a small fraction of 
wearables users. Please note that we demonstrate it in a  scenario 
where the labeled dataset to total dataset size was ~46\%. 
However, in realistic scenarios, it might be a much more smaller proportion. 
Hence, it is pertinent to use an unsupervised method like 
\acttovec\ to harness human activity data from all the users, to improve 
the performance as well as generalization of downstream supervised 
tasks.

The multi-task learning framework boosts the performance across the 
board, exploiting the co-morbidity structure of these multiple disorders, 
underpinning the root cause --- common life-style choices as captured 
partially by the wearable activity signals.

\section{Conclusions} \label{sec:conclusion}
Given the remarkable popularity of wearable devices for human activity tracking, there is a significant potential for personalized automated health-care that can not only reduce health-care costs but also help patients avoid long waiting times. Such a system can potentially alert patients to the risk of an impending health event, and can help in early treatments. Owing to absence of diagnosis data, \eg\ patient EHR, majority of valuable activity data becomes ineffectual. Disorder detection also involves serious generalization issues like skewed distribution and ethnic differences. In such scenarios, an unsupervised representational learning approach can effectively encode common human activity patterns in comparison to task-specific supervised learning approaches that by itself may not generalize well across multiple prediction tasks. 

We model human activity time-series data using an unsupervised representational learning approach that can encode time-series at different granularity levels while modeling local and global activity patterns. 
We train our model on 28,868 days of actigraphy data from 4,124 subjects.
By testing our models on prediction tasks for commonly occurring disorders, we find that day-level granularity preserves the best representations. This is not surprising, since a day is the natural timescale for a full cycle of human activities. Our model, the first task-agnostic representational learning time-series model using simple linear classifiers, beats existing symbolic representation models on several disorder prediction tasks. These symbolic time-series models are computationally expensive, and hard to scale unless an expert feature extraction is performed, while our model learns the representational features automatically, giving better performance on multiple tasks using simple linear classifiers. We further demonstrated that these embeddings can be utilized for pre-training the supervised learning tasks, boosting their performance. 

Co-morbidity occurs among different health disorders owing to common life-style choices, as captured partially in activity patterns. We successfully demonstrate a multi-task learning framework for leveraging the co-morbidity structure, improving the performance on the individual disorder prediction tasks. 

\vspace{-0.5em}
\paragraph{Future Work} Our current \acttovec\ model is not compositional in the sense that it does not combine the representations of lower-level (\eg\ device-generated symbols) units to get representations for the higher-level units (\eg\ hour segments). In future, we would like to investigate compositional structures like convolutional neural network or recurrent neural networks for \acttovec. We would also like to investigate other loss types like ranking loss and reconstruction loss in variational auto-encoders. For the supervised learning, we only used CNNs, in future we also plan to use recurrent architectures for supervised learning tasks. We also plan to work on alternative formulations for multi-task framework to avoid tedious and expensive grid-search for setting task weights. 

\bibliographystyle{ACM-Reference-Format}
\bibliography{references} 

\end{document}